\documentclass[pdflatex,sn-mathphys]{sn-jnl}
\usepackage{latexsym}
\usepackage{graphicx}
\usepackage{amsfonts,amssymb,amsmath}
\usepackage{hyperref}

\jyear{2021}%

\theoremstyle{thmstyleone}%
\theoremstyle{thmstyletwo}%

\theoremstyle{thmstylethree}%

\begin{document}

\title[FPPN: Future Pseudo-LiDAR Frame Prediction for Autonomous Driving]{FPPN: Future Pseudo-LiDAR Frame Prediction for Autonomous Driving}


\author[1,2]{\fnm{Xudong} \sur{Huang}}\email{19120299@bjtu.edu.com}

\author*[1,2]{\fnm{Chunyu} \sur{Lin}}\email{cylin@bjtu.edu.cn}

\author[1,2]{\fnm{Haojie} \sur{Liu}}

\author[1,2]{\fnm{Lang} \sur{Nie}}

\author[1,2]{\fnm{Yao} \sur{Zhao}}

\affil[1]{\orgdiv{With Institute of Information Science}, \orgname{Beijing Jiaotong University}, \orgaddress{\city{Beijing}, \postcode{100044}, \country{China}}}

\affil[2]{ \orgname{Beijing Key Laboratory of Advanced Information Science and Network Technology}, \orgaddress{\city{Beijing}, \postcode{100044}, \country{China}}}


\abstract{LiDAR sensors are widely used in autonomous driving due to the reliable 3D spatial information. However, the data of LiDAR is sparse and the frequency of LiDAR is lower than that of cameras. To generate denser point clouds spatially and temporally, we propose the first future pseudo-LiDAR frame prediction network. Given the consecutive sparse depth maps and RGB images, we first predict a future dense depth map based on dynamic motion information coarsely. To eliminate the errors of optical flow estimation, an inter-frame aggregation module is proposed to fuse the warped depth maps with adaptive weights. Then, we refine the predicted dense depth map using static contextual information. The future pseudo-LiDAR frame can be obtained by converting the predicted dense depth map into corresponding 3D point clouds. Experimental results show that our method outperforms the existing solutions on the popular KITTI benchmark.}




\maketitle

\section{Introduction}\label{sec1}

 Depth information plays a critical role in autonomous driving applications \cite{schneider2017multimodal}, such as object detection \cite{zhao2019contrast} and semantic segmentation \cite{hu2019acnet,kundu2020virtual}. LiDAR sensors provide reliable depth information and are widely used for autonomous driving. However, the point cloud data of LiDAR is sparse in space, and its frequency is rather lower than the popular cameras'. It is meaningful to generate denser point clouds spatially and temporally.

Recently, the combined applications \cite{ku2018joint, krispel2020fuseseg} of point clouds and RGB images have achieved an amazing improvement in computer vision tasks. However, LiDAR sensors have a low frequency(around 10Hz). Therefore, there is a noticeable dislocation between the LiDAR and other sensors such as cameras(around 20Hz) in a multi-sensor system. To increase the frequency of the LiDAR, Liu $et\ al.$ \cite{liu2020plin} propose to interpolate an intermediate frame. However, the interpolation in two consecutive frames is a post-processing operation, which cannot be applied to high-speed practical application scenarios.
\begin{figure}
	\centering
	\includegraphics[width=1\linewidth]{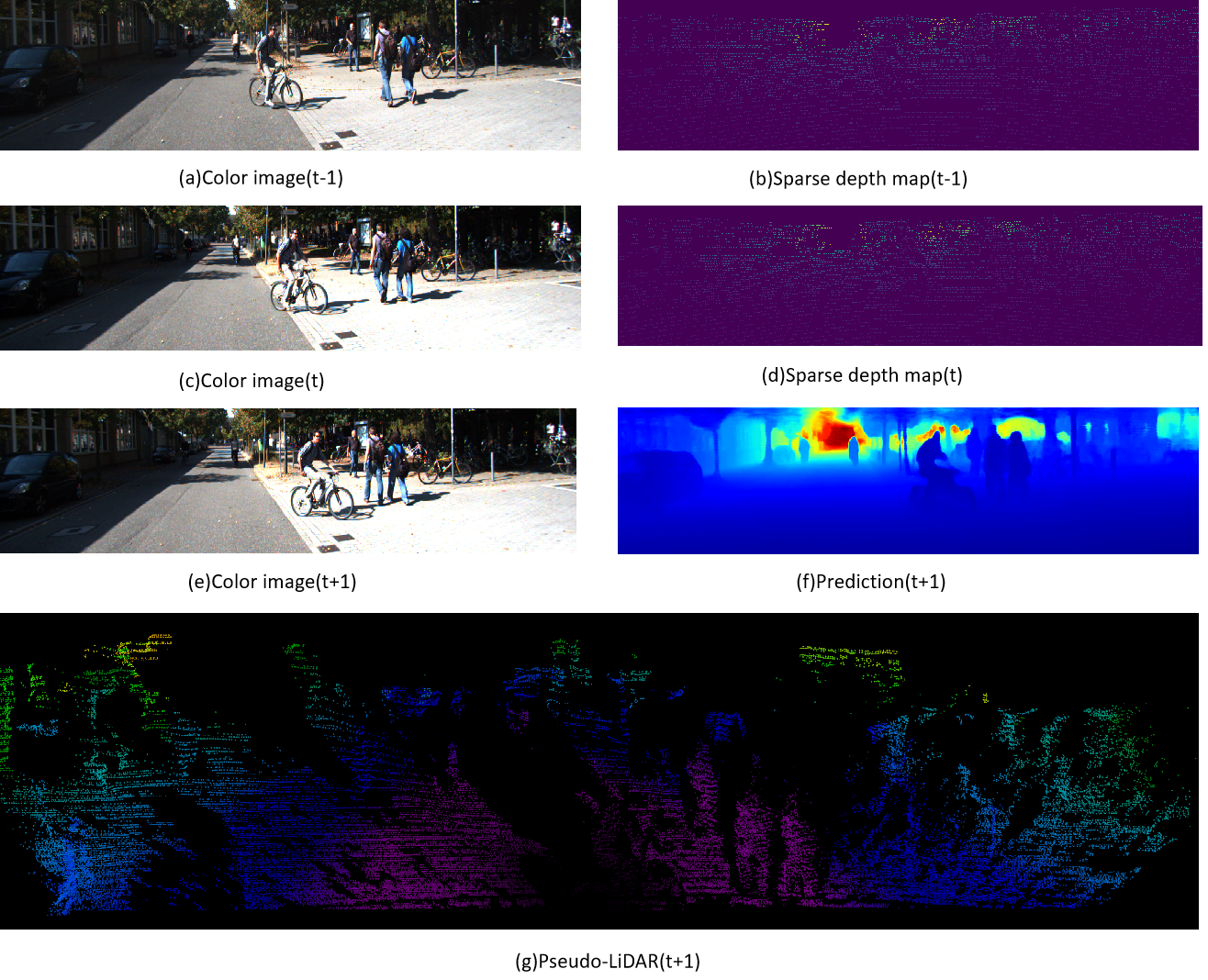}
	\caption[Fig.1.]{The neural network for pseudo-LiDAR prediction in motion scene. Given a sequence of RGB images (a)(c)(e) and two consecutive sparse depth maps (b)(d), the predicted dense depth map is shown as (f). Besides, the final pseudo-LiDAR (g) is obtained by converting the predicted depth map (f).}
	\label{fig:show}
\end{figure}

\begin{figure*}
	\centering
	\includegraphics[width=1\linewidth]{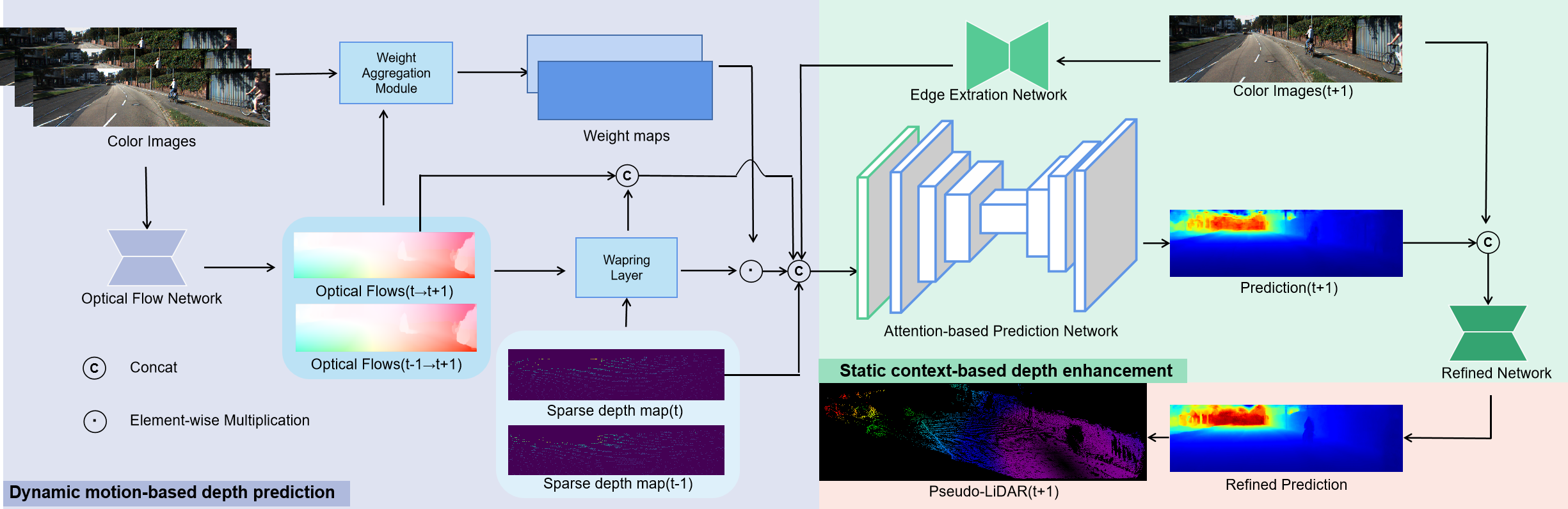}
	\caption[Fig.1.]{Overall Architecture. The whole structure can be divided into two parts: the dynamic motion-based depth prediction module and the static context-based depth enhancement module. In attention-based prediction network, the cuboid with a green border is the attention module.}
	\label{fig:overall}
\end{figure*}
To solve the above problems, we present a future pseudo-LiDAR frame prediction network to generate denser point cloud sequences (Fig.1 (g)) temporally and spatially. The proposed network is composed of a dynamic motion-based depth prediction module and a static context-based depth enhancement module. In the dynamic motion-based depth prediction module, we extract pixel-level motion information such as optical flow to predict the dense depth maps of future frames. To combine the motion relationship of consecutive frames and reduce the errors in optical flow estimation, an adaptive aggregation module is proposed to fuse consecutive warped depth maps. In the static context-based depth enhancement module, we use contextual information such as edge features and RGB images to enhance the predicted results. Furthermore, we feed the predicted dense depth map into the refine network to get more accurate results. The final pseudo-LiDAR is converted from the predicted dense depth map. To our knowledge, this is the first work to predict the future pseudo-LiDAR frame for autonomous driving. Compared with the depth completion methods and the depth interpolation methods, our solution achieves better performance in the depth prediction task on the popular KITTI benchmark \cite{geiger2013vision}.

In summary, our main contributions are concluded as follows:
\begin{itemize}
	\item[$\bullet$]To exploit motion information and contextual information reasonably, we present a dynamic-to-static prediction network to generate the future pseudo-LiDAR which consists of two modules: dynamic motion-based depth prediction and static context-based depth enhancement.
\end{itemize}
\begin{itemize}
	\item[$\bullet$]To fuse warped depth maps appropriately and eliminate the errors in the optical flow estimation, we design an efficient aggregation module to fuse consecutive depth information.
\end{itemize}
\begin{itemize}
	\item[$\bullet$]To our knowledge, this is the first work to predict the future pseudo-LiDAR frame, contributing to generating dense depth map sequences temporally and spatially. 
\end{itemize}

\section{Related work}\label{sec2}

In this section, related works on depth completion, inter-frames interpolation, and video prediction will be described.

\subsection{Depth completion}

The depth completion task aims to obtain an accurate dense depth map by completing a sparse depth map. According to whether there is an RGB image for guidance, previous methods can be roughly divided into two categories: non-guided depth completion \cite{uhrig2017sparsity, eldesokey2020uncertainty} and image-guided depth completion \cite{liu2013guided, qiu2019deeplidar, park2014high}. The early method \cite{herrera2013depth} complements sparse depth maps by optimizing the energy function. \cite{ku2018defense} uses the hand-crafted features to convert sparse depth maps into dense. However, the use of hand-craft features leads to poor performance. More recently, deep learning-based methods \cite{qiu2019deeplidar,van2019sparse,ma2019self} for depth completion outperforming traditional methods by a wide margin. More specifically, \cite{xu2019depth} models the geometric constraints between depth and surface normal to improve the robustness against noise. \cite{park2020non} estimates non-local neighbors and their affinities of each pixel to concentrates on relevant non-local neighbors during completion. To make better use of images, \cite{tang2020learning} designs a novel module to predict kernel weights from the guidance image, which extracts the depth image features better. Inspired by these, we use a neural network for transforming a sparse depth map into dense.

\subsection{Inter-frames interpolation}

Video interpolation aims to generate non-existent frames from adjacent frames in two-dimension. It can generate temporally high-quality video sequences. \cite{meyer2015phase} flows pixel values from existing ones to synthesize video frames. \cite{Jiang_2018_CVPR} proposes an end-to-end convolutional neural network to jointly model the motion interpretation and occlusion reasoning. Compared with video frame interpolation, three-dimensional frame interpolation is more challenging. To solve the low-frequency problem of the LiDAR, \cite{liu2020plin} utilizes a learning-based method to predict bi-directional optical flow to estimate the motion between consecutive frames. The two input frames are further warped and generate the intermediate depth frames through a coarse-to-fine cascade structure. Then, the pseudo-LiDAR \cite{wang2019pseudo} can be converted from the depth map using camera parameters. Another solution is to interpolate an intermediate point cloud straightly using two consecutive point cloud frames \cite{lu2021pointinet}. It upsamples low frame rate LiDAR point cloud streams to higher frame-rate. However, interpolation in two consecutive frames is only a post-processing work, which cannot be applied to high-speed practical application scenarios.

\subsection{Frame prediction}

Compared to inter-frame interpolation, video prediction \cite{lotter2016deep} is a future frame interpolation task that combines spatio-temporal sequence information to get the future frame. Predicting the future frame is a multi-modal task as there are multiple possibilities. The key to predicting the precise future frame is taking advantage of the temporal sequence information. Currently, a large number of frame prediction methods are based on GAN \cite{tulyakov2018mocogan,vondrick2016generating,liang2017dual} or VAE \cite{walker2016uncertain}. For unsupervised methods, \cite{finn2016unsupervised} proposes an action-conditioned model that explicitly models pixel motion by predicting a distribution over pixel motion from previous frames. Compared to the success of video frame prediction, frame prediction of depth maps has not been well explored.

\section{Method}\label{sec3}

\subsection{Overall architecture}

Given the past two frames of sparse depth maps \emph{d$_{t-1}$,d$_{t}$} and three consecutive RGB images \emph{$I_{t-1},I_{t},I_{t+1}$}, our network predicts the future dense depth map \emph{$\hat{D}_{t+1}$}. As shown in Fig.\ref{fig:overall}, our network can be divided into two parts: dynamic motion-based depth prediction and static context-based depth enhancement. In the dynamic motion-based depth prediction part, we first get a sparse predicted depth map by estimating the mutual optical flows between frames. Then, a dense prediction is obtained from an attention-based prediction network. To reduce errors in optical flow estimation, we propose an adaptive depth map aggregation module to fuse consecutive depth information. Since the dynamic motion-based depth prediction module neglects the static contextual details, we design the static context-based depth enhancement module to enhance the prediction results. 

\subsection{Dynamic motion-based depth prediction}

\subsubsection{Depth map warping}\label{warp}

Compared with predicting the future frame from a single frame, consecutive frames contain abundant motion details. Considering that there is a motion relationship between consecutive frames in a motion scene, we predict the approximate sparse depth map using the optical flow that contains rich motion information. We use the lightweight framework Liteflownet \cite{hui2018liteflownet} to estimate optical flow. Given three consecutive RGB images \emph{I$_{t-1}$,I$_{t}$,I$_{t+1}$}, optical flows can be obtained by:
\begin{equation}\label{1}
	\begin{aligned}
		F_{t+1\rightarrow t}&=\mathcal{F}(I_{t+1},I_{t})\\
		F_{t+1\rightarrow t-1}&=\mathcal{F}(I_{t+1},I_{t-1})
	\end{aligned}
\end{equation}	
where $\mathcal{F}$ is the neural network for optical flow estimation, $F_{t+1\rightarrow t-1}$ is the optical flow from $I_{t+1}$ to $I_{t-1}$ and $F_{t+1\rightarrow t}$ is the optical flow from $I_{t+1}$ to $I_{t}$. Suppose that the estimated optical flow is an approximate representation of the inter-frame motion information, the sparse predicted depth map $d_{t+1}$ can be expressed as:
\begin{equation}\label{2}
	\begin{aligned}
		d_{t\rightarrow t+1}&=\mathcal{W}(d_{t},F_{t+1\rightarrow t})\\
		d_{t-1\rightarrow t+1}&=\mathcal{W}(d_{t-1},F_{t+1\rightarrow t-1})
	\end{aligned}
\end{equation}	
where $\mathcal{W(,)}$ is the back warping operation. $d_{t\rightarrow t+1}$ and $d_{t-1\rightarrow t+1}$ represent the predicted sparse depth maps, which are warped from $d_{t}$ and $d_{t-1}$.
\begin{figure*}
	\centering
	\includegraphics[width=1\linewidth]{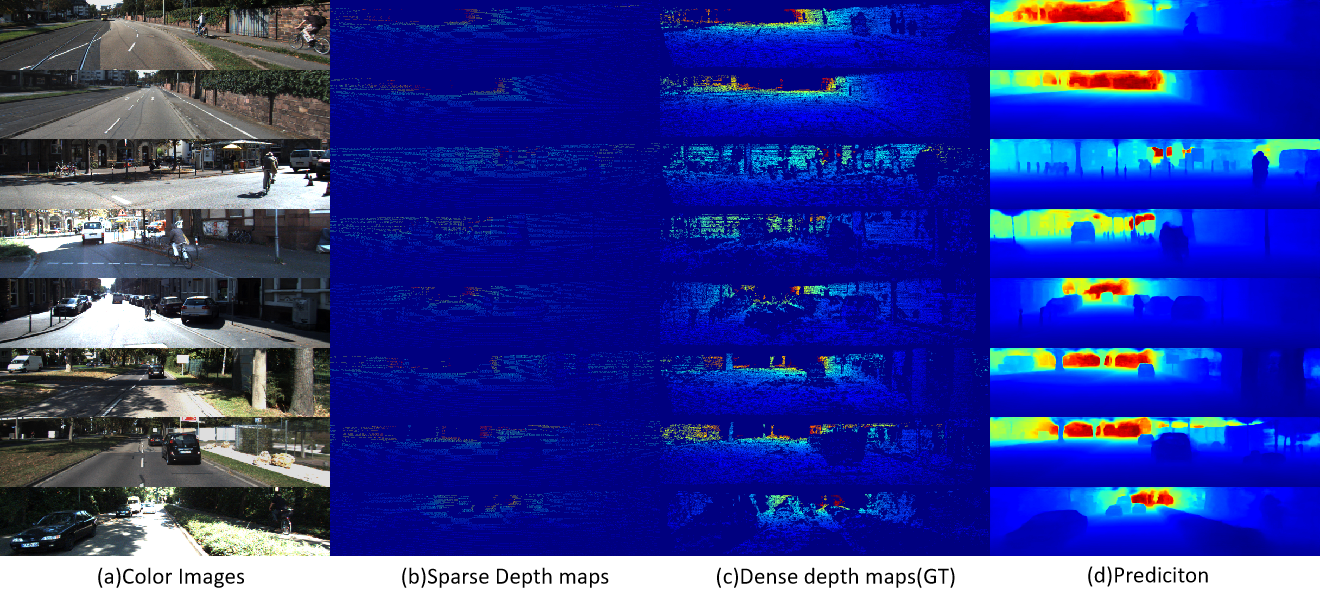}
	\caption[Fig.3.]{Prediction results(d) at t+1. We show t+1-th RGB images(a), sparse depth maps(b) and dense ground truth(c). Our method can generate dense depth maps and is closer to ground truth.}
	\label{fig:result}
\end{figure*}

\subsubsection{Adaptive depth map aggregation}

Once we obtain the warped sparse depth maps from Eq.\ref{2}, the key to predicting the precise future frame is how to take advantage of the temporal sequence information. A straightforward solution is to combine warped depth maps directly by extracting features individually and then feeding them into the prediction network. However, this rough approach regards $d_{t\rightarrow t+1}$ and $d_{t-1\rightarrow t+1}$  as equally important, ignoring their chronological order. Besides, this solution neglects the depth errors caused by the optical flow estimation. To solve the above problems, we propose an aggregation module that can fuse the warped depth maps with adaptive weights and enhance the robustness of our model to optical flow estimation errors. Specifically, the weight maps can be obtained by calculating the similarity matrix between warped RGB images and actual RGB images. Then, we use the weight maps to guide the depth maps aggregation, rejecting the unreliable depth values.

To implement this module, we first obtain the warped RGB images {$I_{t\rightarrow t+1}$,$I_{t-1\rightarrow t+1}$} using the estimated optical flow. Then L2-normalization is performed on the warped image. Finally, we calculate the cosine similarity matrix \cite{luo2018cosine} between the warped images and feed the matrix into the softmax function to obtain the weight maps. More specifically, the aggregation weight at pixel $(i,j)$ can be expressed as:
\begin{equation}\label{3}
	\begin{aligned}
        w_{t-1}^{ij}=&softmax(\frac
		{ I^{ij}_{t-1\rightarrow t+1 }\cdot  I^{ij}_{t+1}}
		{\vert I^{ij}_{t-1\rightarrow t+1 }\vert  \vert I^{ij}_{t+1}\vert})\\
		w_{t}^{ij}=&softmax(\frac
		{ I^{ij}_{t\rightarrow t+1 }\cdot  I^{ij}_{t+1}}
		{\vert I^{ij}_{t\rightarrow t+1 }\vert  \vert I^{ij}_{t+1}\vert})
	\end{aligned}
\end{equation}
$I^{ij}$ denotes the vector of pixel(i,j).Then, the aggregated sparse depth map $d_{t+1}^{A}$ can be expressed as:

\begin{equation}\label{eq:aggregation}
	\begin{aligned}
		d_{t+1}^{A}=w_{t-1}\cdot d_{t-1\rightarrow t+1}+w_{t}\cdot d_{t\rightarrow t+1}
	\end{aligned}
\end{equation}

\subsubsection{Predict depth map}

We formulate the depth map prediction problem as a pixel-level regression problem. To implement it, the proposed network follows an encoder-decoder structure \cite{ronneberger2015u}. In the encoder part, We use four consecutive ResNet-34 \cite{he2016deep} and a convolutional layer with 3×3 filter to extract features. In the decoder part, five deconvolutional layers and a convolutional layer with a 1×1 filter are designed to generate the output.To prevent the loss and forgetting of the information during the convolution process, feature maps of the same size are connected by skip connections. To make better use of the motion information, we add the optical flow and the aggregated warped depth map to the input of the prediction network. All inputs are fed into initial convolutions respectively. Except for the last layer, each convolutional layer is connected to batch normalization and ReLU. 

Therefore, the input of prediction network in motion-based part includes the t-th sparse depth maps, the estimated
optical flow from t-th to t+1-th, the warped sparse depth maps and the aggregated sparse depth map. The predicted dense depth map can be expressed as:
\begin{equation}\label{4}
	\begin{aligned}
		{D}_{t+1}^{C}=\mathcal{P}(d_{t},F_{t+1\rightarrow t},d_{t\rightarrow t+1},{d}_{t+1}^{A})
	\end{aligned}
\end{equation}
where $\mathcal{P}$ denotes the depth map prediction network and \emph{${D}_{t+1}^{C}$} denotes the predicted dense depth map.

\subsection{Static context-based depth enhancement }

To generate an accurate and dense depth map, we design a static context-based depth enhancement module to combine contextual information. The texture feature can guide the network to pay more attention to the saliency objects, containing more clear structure and details. We first get the edge features \cite{canny1986computational} extracted from $I_{t+1}$ and feed them into initial convolution. Then, the obtained edge features are concatenated with input for prediction in Eq.\ref{4}. The predicted dense depth map can be updated as:
\begin{equation}\label{5}
	\begin{aligned}
		{D}_{t+1}^{C}=\mathcal{P}^{'}(d_{t},F_{t+1\rightarrow t},d_{t\rightarrow t+1},d_{t+1}^{A},\mathcal{E}_{t+1})
	\end{aligned}
\end{equation}
where $\mathcal{E}_{t+1}$ is the t+1-th edge feature and $\mathcal{P}^{'}$ the prediction network that joins edge information. Note that all inputs are fed into initial convolutions respectively.

Since our input has a lot of feature maps and is sensitive to spatial information, we use the attention module Convolutional Block Attention Module(CBAM) \cite{woo2018cbam} to increase the use of static spatial information and multi-scale feature maps. More specifically, we feed the input feature $F$ into the channel-attention module and spatial-attention module to get the corresponding attention maps. Then the final feature $F'$ is generated by multiplying the attention maps with the input feature. The final feature map $F'$ obtained from the CBAM module can be expressed as:
\begin{equation}\label{6}
	\begin{aligned}
		F_{c}&= M_{c}(F)\otimes F\\
		F'&=M_{s}(F_{c})\otimes F_{c}
	\end{aligned}
\end{equation}
where $F$ denotes the concated intermediate feature maps of the prediction network, $\otimes$ denotes element-wise multiplication, $M_{c}(\cdot)$ denotes the channel attention maps, $M_{s}(\cdot)$ denotes the spatial attention maps. 

To obtain more accurate prediction results, we design a refined module on the guidance of the future RGB image. We take U-net \cite{ronneberger2015u} as the refinement network. Specifically, the refined network contains five convolutions in the encoder and five deconvolutions in the decoder. The batch normalization layer and ReLU are connected after each convolutional layer. The output of each encoding convolution will be connected to the corresponding decoding layers via skip connections. We take the predicted results and t+1-th RGB images as inputs. The final predicted dense depth map $\hat{D_{t+1}}$ can be expressed as:
\begin{equation}\label{7}
	\begin{aligned}
		\hat{D}_{t+1}=W(D_{t+1}^{C},I_{t+1})
	\end{aligned}
\end{equation}
\begin{figure*}
	\centering
	\includegraphics[width=1\linewidth]{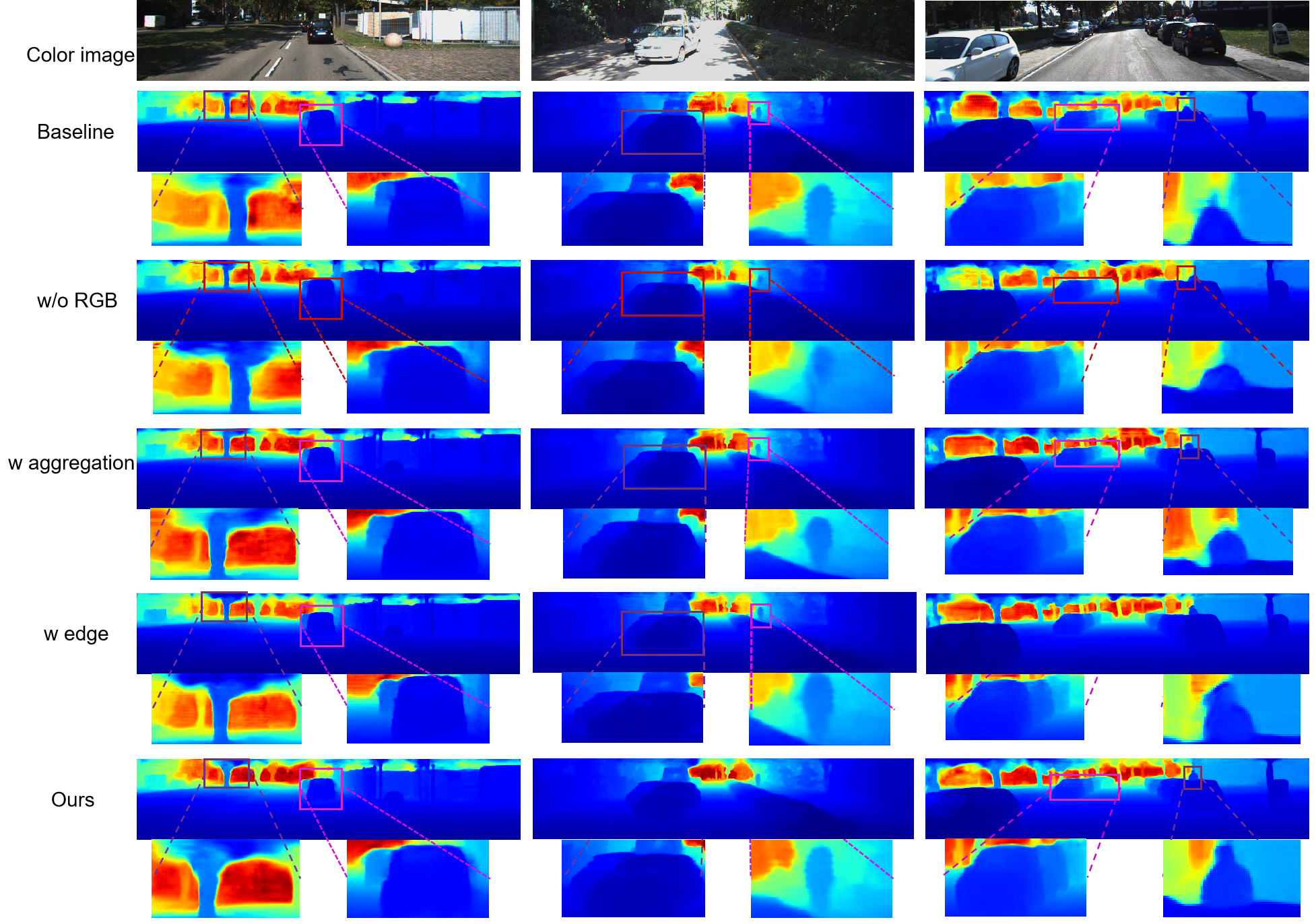}
	\caption[Fig.4.]{Comparison of different components. We visualize the predicted results with different components, the complete network generates not only more accurate results, but also clearer boundaries.}
	\label{fig:ablation}
\end{figure*}
\subsection{Pseudo-LiDAR generation}
Once the predicted dense depth map is generated, the pseudo-LiDAR \cite{wang2019pseudo} can be converted from the depth map using camera parameters. According to the pinhole camera model principle, we can derive the 3D location $(x, y, z)$ of each pixel $(u, v, d)$ where $d$ represents the depth value. The corresponding relationship is described as follows:

\begin{equation}\label{z}
	\begin{aligned}
		z=d(u, v),
	\end{aligned}
\end{equation}
\begin{equation}\label{x}
	\begin{aligned}
		x=\frac{(u-c_{U})\times z}{f_{U}},
	\end{aligned}
\end{equation}
\begin{equation}\label{y}
	\begin{aligned}
		y=\frac{(v-c_{V})\times z}{f_{V}},
	\end{aligned}
\end{equation}
where $(c_{U}, c_{V})$ is the pixel location corresponding to the center of the camera, and $f_{V}$ and $f_{U}$ are the vertical focal length and horizontal focal length of the camera.

By mapping all the pixels of dense depth map into 3D coordinates, we obtain a point cloud {($x^{(n)}$,$ y^{(n)}$, $z^{(n)}$)}$^{N}_{n=1}$, where $N$ is the point count. Since this point cloud is converted from the depth map using camera parameters, we refer to the point cloud as a pseudo-LiDAR.
\subsection{Loss}
The final output of our network is a dense depth map while the ground truth is a sparse depth map, so we use mask L2 loss as the loss function of the network. More specifically:
\begin{equation}\label{8}
	\begin{aligned}
		\mathcal{L}_{mask}= \left\|(x-\hat{x})\odot \mathfrak{1}(\hat{x}>0) \right\|_2^{2}
	\end{aligned}
\end{equation}
where $\|\cdot\|_2^{2}$ denotes the L2-norm, $\mathfrak{1}(\cdot)$ denotes the mask of sparse ground depth and $\odot$ is an element-wise multiplication. Since our module can be divided into dynamic motion-based depth prediction and static context-based depth enhancement, our loss function can be obtained by linearly combining the losses of the two modules.
Our final loss function can be expressed as:
\begin{equation}\label{9}
	\begin{aligned}
		\mathcal{L}= \lambda_{1}\mathcal{L}_{mask}(D_{t+1}^{C},D_{t+1}^{GT})+\lambda_{2}\mathcal{L}_{mask}(\hat{D}_{t+1},D_{t+1}^{GT})
	\end{aligned}
\end{equation}
where $D_{t+1}^{GT}$ denotes the ground truth at t+1-th. $\lambda_{1}$ and $\lambda_{2}$ are the hyper-parameters set to 0.1 and 0.9 based on experience. 

\section{Results}\label{sec4}

\subsection{Experiment setting}

\subsubsection{Dataset}

We evaluate our method on one of the largest depth completion and depth prediction datasets for outdoor scene prediction: KITTI depth completion datasets \cite{uhrig2017sparsity}. It contains over 93 thousand depth maps with corresponding raw LiDAR scans and RGB images. And it is aligned with the ``raw data'' of the KITTI dataset \cite{geiger2013vision} which comprises Velodyne point clouds, calibration, etc. Since this is a prediction task, we will remove the data with relatively small movements. Finally, we choose 50,390 data as the training set, 6,852 as the validation set, and 1,000 as the test set.

\subsubsection{Training details}
The point cloud is relatively sparse at the top and can't provide any depth information. Therefore, we crop the input with size 1216×256 from the top. We use the Adam optimizer \cite{kingma2014adam} with an initial learning rate of 10$^{-5}$ and decrease to half every 5 epochs. Momentum is 0.9. In all cases, we perform data augmentation by flipping vertical. Finally, we set batch size as 1 and train the model on an NVIDIA RTX 2080Ti GPU.

\subsubsection{Evaluation metrics}
Since our output is a dense depth map, we can use the evaluation metrics of depth completion to evaluate the quality of the generated prediction. Four metrics: Root mean squared error(RMSE), Mean absolute error(MAE), root mean squared error of the inverse depth(iRMSE), and mean absolute error of the inverse depth(iMAE) are used to evaluate the KITTI benchmark. We use RMSE as the primary metric because it reflects the accuracy of the depth map and penalizes larger errors. These four evaluation metrics can be expressed as:
 \begin{equation}\label{RMSE}
 	\begin{aligned}
 		RMSE=\sqrt{\frac{1}{m}\sum_{i=1}^M(d_{pred}^{(i)}-d_{gt}^{(i)})}
 	\end{aligned}
 \end{equation}
 \begin{equation}\label{MAE}
	\begin{aligned}
		MAE=\frac{1}{m}\sum_{i=1}^M\vert(d_{pred}^{(i)}-d_{gt}^{(i)})\vert
	\end{aligned}
\end{equation}
 \begin{equation}\label{iMAE}
	\begin{aligned}
		iMAE=\frac{1}{m}\sum_{i=1}^M\vert(\frac{1}{d_{pred}^{(i)}}-\frac{1}{d_{gt}^{(i)}})\vert
	\end{aligned}
\end{equation}
 \begin{equation}\label{iRMSE}
	\begin{aligned}
		iRMSE=\sqrt{\frac{1}{m}\sum_{i=1}^M(\frac{1}{d_{pred}^{(i)}}-\frac{1}{d_{gt}^{(i)}})}
	\end{aligned}
\end{equation}

\subsection{Comparison results}

\subsubsection{Compared with depth completion methods}

We show the comparison results with the depth completion methods in Table \ref{tab:PE}. PENet \cite{hu2021penet} achieved state-of-the-art results and ACMNet \cite{zhao2021adaptive} ranked second for depth completion task on the KITTI dataset. The depth completion task aims to densify the sparse depth map. In contrast, due to the use of motion information in the dynamic motion-based prediction part, our method has a good performance in the dense depth map prediction task. The results also show that our method has a lower root mean squared error(RMSE) and mean absolute error(MAE) in Table \ref{tab:PE}.

\subsubsection{Compared with dense depth interpolation methods}

Similar to our work, the dense depth map interpolation aims to generate dense point cloud sequences by interpolation in the consecutive sparse depth map. However, the interpolation is only a post-processing work, which cannot be applied to high-speed practical application scenarios. To prove that our method can increase the frequency of LiDAR sensors and use motion information to satisfy the needs of real-time applications, we compare our approch with the dense depth map interpolation method. Table \ref{tab:PLIN} shows the results that our algorithm is specially designed for the dense depth prediction. Moreover, our method is jointly guided by dynamic and static information, achieving the best performance.

\begin{table}
	\centering
	\caption{Quantitative comparison with depth completion methods.}
	\begin{tabular}{|l|c|c|c|c|}\hline
		Method&RMSE$\downarrow$&MAE$\downarrow$&iRMSE$\downarrow$&iMAE$\downarrow$\\\hline
		PENet \cite{hu2021penet}&1571.59&598.63&5.89&2.76\\
		ACMNet \cite{zhao2021adaptive}&2247.05&913.05&9.42&4.50\\
		ours&\textbf{1214.96}&518.34&6.52&3.32\\\hline
	\end{tabular}
	
	\label{tab:PE}
\end{table}

\begin{table}
	\centering
	\caption{Quantitative comparison with depth interpolation methods.}
	\begin{tabular}{|l|c|c|c|c|}\hline
		Method&RMSE$\downarrow$&MAE$\downarrow$&iRMSE$\downarrow$&iMAE$\downarrow$\\\hline
		PLIN \cite{liu2020plin}&1302.49&558.02&6.43&3.34\\
		PLIN\_V2 \cite{liu2021pseudo}&1412.21&511.69&6.01&2.59\\
		ours&\textbf{1214.96}&518.34&6.52&3.32\\\hline
	\end{tabular}
	
	\label{tab:PLIN}
\end{table}

\subsection{Ablation study and analysis}

To evaluate the performance of each component of the network, ablation studies are conducted and the results are presented in Table \ref{tab:ablation}. Besides, to visibly compare the impact of different components, the ablation experimental results are shown in Fig.\ref{fig:ablation}.

\noindent\textbf{Baseline.} To simply build a prediction task using motion information, we feed the previous depth map $d_{t}$, the warped depth maps $d_{t\rightarrow t+1}$ and $d_{t-1\rightarrow t+1}$ into prediction network with the refinement module. 

\noindent\textbf{Refinement module.}It is worth noting that there is a significant improvement using the refinement network. The RGB image has abundant contextual information and can guide the network to pay more attention to the saliency objects, which contain more clear structure and details. 

\noindent\textbf{Adaptive aggregation module.} When adaptive aggregation input is used, the RMSE decreases because the acceleration information is better utilized, not just the motion information. The error in optical flow estimation leads to a ripple in the predicted results. The aggregation module reduces the bias in motion estimation by aggregating consecutive frames which can be shown by the pink boxes in the figure.

\noindent\textbf{Edge information.} The addition of edge information also has a positive impact on the results. It is evident that the boundary of the prediction result using edge information is more apparent in purple boxes.

\noindent\textbf{Attention module.} When the attention mechanism is used, the RMSE is severely degraded because the attention module tries to find the focal region from a larger number of feature maps and spatial domains. 

\begin{table}
	\centering
	\caption{Ablation study on the selected validation set of the KITTI depth completion dataset.}
	\begin{tabular}{|l|c|c|c|c|c|}\hline
		Setting&RMSE$\downarrow$ &MAE$\downarrow$ &iRMSE$\downarrow$ &iMAE$\downarrow$  \\\hline
		baseline&1323.99&612.03&7.71&4.24\\
		\quad$-$ rgb&1441.40&601.46&7.56&3.56\\
		\quad$+$ aggregation&1288.67&584.65&7.22&3.78\\
		\quad+ edge&1297.55&518.65&\textbf{6.25}&\textbf{3.08}\\
		\quad+ attention&1292.21&572.92&7.02&3.71\\
		ours&\textbf{1214.96}&\textbf{518.34}&6.52&3.32\\\hline
	\end{tabular}
	
	\label{tab:ablation}
\end{table}

\section{Conclusion}\label{sec5}

This paper presents a method for depth map prediction with two consecutive depth maps and RGB image sequences. We first use the dynamic motion-based prediction module to obtain a predicted depth map. Then we propose a static context-based augmentation for refinement to get more accurate predicted dense depth maps. The final LiDAR is obtained by converting the predicted depth map. Specifically, we recover motion information using the estimated optical flow and then use the adaptive aggregation module to get a more accurate motion scene. To get more accurate prediction results, we combine context-based information: edges and RGB images that contain rich contextual information. Finally, the attention module is used to focus on the region of interest. We evaluate our model on the KITTI benchmark, which predicts the future frame in motion scenes and verifies the validity of the proposed method.

\backmatter


\bibliography{sn-bibliography}


\end{document}